\pdfoutput=1
\documentclass{bmvc2k}
\usepackage{amsmath}
\usepackage{amssymb}
\usepackage{amsthm}
\usepackage{multirow}
\usepackage{makecell}
\usepackage{diagbox}
\usepackage{booktabs}

\title{Masked Vision-Language Transformers for Scene Text Recognition}

\addauthor{Jie Wu}{wu.jie@westone.com.cn}{1}
\addauthor{Ying Peng}{peng.ying06059@westone.com.cn}{1}
\addauthor{Shengming Zhang}{shmizhang@gmail.com}{1}
\addauthor{Weigang Qi}{qi.weigang@westone.com.cn}{1}
\addauthor{Jian Zhang}{zhang.jian06086@westone.com.cn}{1}

\addinstitution{
 Westone Information Industry INC.\\
 Chengdu, China
}

\runninghead{Wu et al.}{Masked Vision-Language Transformers for STR}

\begin{document}

\maketitle
\begin{abstract}
Scene text recognition (STR) enables computers to recognize and read the text in various real-world scenes. Recent STR models benefit from taking linguistic information in addition to visual cues into consideration. We propose a novel \textbf{M}asked \textbf{V}ision-\textbf{L}anguage \textbf{T}ransformers (MVLT) to capture both the explicit and the implicit linguistic information. Our encoder is a Vision Transformer, and our decoder is a multi-modal Transformer. MVLT is trained in two stages: in the first stage, we design a STR-tailored pretraining method based on a masking strategy; in the second stage, we fine-tune our model and adopt an iterative correction method to improve the performance. MVLT attains superior results compared to state-of-the-art STR models on several benchmarks. Our code and model are available at \url{https://github.com/onealwj/MVLT}.
\end{abstract}

\section{Introduction}
\label{Introduction}
Scene text recognition (STR) aims to read text from natural scenes, which is helpful in many practical artificial intelligence applications such as autonomous driving, instant translation, and natural scene understanding. STR has been studied extensively in the past two decades, however, the performance of which still struggles under scenarios with unpromising illuminations, occluded characters, complex deformations, etc. 
Recent studies \cite{jaderberg2015deep, jaderberg2014deep,  DBLP:conf/cvpr/YuLZLHLD20, DBLP:conf/cvpr/FangXWM021, tang2021visual, DBLP:conf/ijcai/DuCJYZLDJ22, DBLP:conf/iccv/WangXFWZZ21} have made progress in dealing with such challenges by introducing textual semantics information except for visual cues. 
These language-aware methods are typically divided into two kinds: explicitly building an extra language model or implicitly extracting textual semantics from visual cues.
The former \cite{jaderberg2015deep, jaderberg2014deep, DBLP:conf/cvpr/YuLZLHLD20, DBLP:conf/cvpr/FangXWM021} rely on language models such as n-grams \cite{jaderberg2015deep} 
or attention-based neural networks \cite{DBLP:conf/cvpr/YuLZLHLD20} to predict word-level text. 
The latter \cite{tang2021visual, DBLP:conf/ijcai/DuCJYZLDJ22, DBLP:conf/iccv/WangXFWZZ21} attempt to guide the model to catch textual semantics according to visual context without using additional language models. For instance,  \cite{DBLP:conf/iccv/WangXFWZZ21} exploits a character-level occluded strategy to make the visual model learn linguistic information along with visual features. 
Despite the effectiveness of both ways, each of them only captures textual semantics from a single orientation. Inspired by their success, we propose a method to learn both the explicit and the implicit textual semantics, promoting the ability of STR.

We propose a \textbf{M}asked \textbf{V}ision-\textbf{L}anguage \textbf{T}ransformers (MVLT) for STR, with a Vision Transformer (ViT) \cite{DBLP:conf/iclr/DosovitskiyB0WZ21} encoder and a creatively designed multi-modal Transformer \cite{DBLP:conf/iclr/SuZCLLWD20, DBLP:conf/aaai/0010TYSTW021,li2019visualbert} decoder.  The potential of Transformers ~\cite{DBLP:conf/nips/VaswaniSPUJGKP17} has been proved in the area of NLP and CV, while the pretraining and fine-tuning pipelines of Transformer-based models further boost the performance of down-streaming tasks. To this end, we adopt a two-stage training strategy to train the model as follows:

In the first stage, we pretrain the model by borrowing the idea from masked autoencoders (MAE) \cite{he2022masked}. MAE splits an image into several patches, randomly masks a percentage of them, and learns to reconstruct the masked ones. Different from MAE, our MVLT recognizes scene text  in addition to reconstructing the masked patches.
To recognize text, we seek to use linguistic information to assist the visual cues. 
Motivated by multi-modal Transformers, which combine image regions with language semantics through a Transformer and learn the interaction between the two, we build a multi-modal Transformer decoder to bring linguistic information into our model. For the input of our multi-modal decoder, the visual part consists of encoded patches and mask tokens; the textual part comes from the ground-truth text label of the corresponding image and  is formed into a sequence of character embeddings. 
We build two sub-decoders to model explicit and implicit textual semantics, respectively.
Similar to MLM in VisualBERT \cite{li2019visualbert}, which masks some tokens in the textual input and predicts them according to the visual and textual features, for one sub-decoder, we mask a part of characters in the ground-truth text label and learn to predict the masked characters. The visible characters serve as word-level textual cues, explicitly guiding the model to learn linguistic knowledge. 
To endow the model with the ability to predict the correct word-level text even without explicit textual input, which is closer to the application scenario of STR, for  another sub-decoder, we mask all input characters. The model is now left  with only visual cues and thus is pushed to learn linguistic information implicitly without an additional language model.
It is worth noticing that the two sub-decoders share parameters during training for efficiency.
Furthermore, as an extension of our MVLT, we propose a simple but efficient method to use real-world unlabeled data together with labeled one in pretraining.

In the second stage, we fine-tune the model, where the encoder takes the unmasked scene text image as input, and the decoder outputs the predicted text. To better use the pretrained knowledge, different from MAE or ViTSTR \cite{DBLP:conf/icdar/Atienza21}, which only fine-tune on the pretrained encoder, we fine-tune on both the encoder and the decoder. Meanwhile, we propose an iterative correction method to gradually modify the predicted text during iterations. In each iteration, the input of the decoder consists of the image feature that is output from encoder, and the text feature that is output from the previous iteration.

Our contributions are three-folds: 
(1) we propose a novel language-aware model for STR, allowing for learning both the explicit and the implicit semantics, which gets superior accuracy compared to previous works;
(2) we design a masking-based strategy in pretraining the model and exploit an iterative correction method to correct text predictions in the fine-tuning stage.
(3) we propose a method to further promote accuracy by leveraging labeled and unlabeled training data. 


\section{Related Work}
\label{Related_Work}
One of the classical ways
for STR is to extract visual features based on CNN, use RNN to perform sequence labeling, and adopt Connectionist Temporal Classification (CTC) \cite{DBLP:conf/icml/GravesFGS06} as a loss function \cite{DBLP:journals/pami/ShiBY17, DBLP:conf/aaai/HeH0LT16}. Recently, GTC \cite{DBLP:conf/aaai/HuCHYL20} optimized the CTC-based method by using a graph convolutional network (GCN) \cite{DBLP:conf/iclr/KipfW17} to learn the local correlations of features. 
Some works attempt to rectify the irregular text images \cite{DBLP:journals/pami/ShiYWLYB19, DBLP:conf/cvpr/ZhanL19, DBLP:conf/iccv/YangGLHBBYB19}. For example, ASTER \cite{DBLP:journals/pami/ShiYWLYB19} adopts Thin-Plate-Spline (TPS) \cite{DBLP:journals/pami/Bookstein89} transformation, and ScRN \cite{DBLP:conf/iccv/YangGLHBBYB19} adds symmetrical constraints in addition to the TPS. Recent works \cite{jaderberg2014deep, DBLP:conf/cvpr/QiaoZYZ020, DBLP:conf/aaai/00130SZ19, DBLP:conf/cvpr/YanPXY21, DBLP:conf/iccv/BhuniaSKGCS21, DBLP:conf/icdar/ShengC019} brought insightful ideas in dealing with challenging scenarios (occlusion, noise, etc.) by building language-aware models.

{\bf Language-aware methods.} 
Some methods \cite{DBLP:conf/cvpr/YuLZLHLD20, jaderberg2015deep, DBLP:conf/cvpr/FangXWM021, DBLP:conf/cvpr/QiaoZYZ020} rely on external language models to extract the semantics. For example, SRN \cite{DBLP:conf/cvpr/YuLZLHLD20} builds a global semantic reasoning module, which learns semantics based on the predicted text from the visual model. This method is upgraded in ABINet \cite{DBLP:conf/cvpr/FangXWM021}, which develops a bidirectional cloze network to make better use of bidirectional linguistic information, and utilizes an iterative correction for the language model. Some other works \cite{tang2021visual, DBLP:conf/ijcai/DuCJYZLDJ22, DBLP:conf/iccv/WangXFWZZ21} implicitly learn semantics without using language models. For example, SVTR \cite{DBLP:conf/ijcai/DuCJYZLDJ22} recognizes both the characters and the inter-character long-term dependence in a single visual model by using local and global mixing blocks. VST \cite{tang2021visual} extracts semantics from a visual feature map and performs a visual-semantic interaction using an alignment module. Most recently, some language-aware models explore more possibilities to boost the performance by considering spatial context \cite{DBLP:conf/aaai/HeC0LHWD22}, adding real-world images to train the model in a semi-supervised way \cite{DBLP:journals/corr/abs-2205-03873}, or developing a re-ranking method to get a better candidate output \cite{DBLP:conf/accv/SabirMP18}, while our work takes a step by extracting both the explicit and the implicit semantics. 

{\bf Transformer-based STR.} Recently, Transformer has shown its effectiveness in STR models \cite{DBLP:conf/cvpr/YuLZLHLD20, DBLP:conf/cvpr/FangXWM021, DBLP:conf/iccv/WangXFWZZ21, DBLP:conf/mm/QiaoZWWZJWW21, DBLP:conf/cvpr/YanPXY21, DBLP:conf/icdar/ShengC019}. 
For instance,  PIMNet \cite{DBLP:conf/mm/QiaoZWWZJWW21} builds a bi-directional Transformer-based parallel decoder to iteratively capture context information.
Lately, based on 
ViT structure \cite{DBLP:conf/iclr/DosovitskiyB0WZ21}, ViTSTR \cite{DBLP:conf/icdar/Atienza21} uses a ViT encoder to perform STR without using the decoder. Additionally, ViTSTR is initialized by pretrained parameters of DeiT \cite{DBLP:conf/icml/TouvronCDMSJ21}. We also apply a ViT encoder in our model. However, unlike ViTSTR, we take the decoder into our architecture and propose a new pretraining method to better fit the STR task.

\section{Methodology}
\label{Method}
\begin{figure}[t]
    \centering
	\includegraphics[width=0.97\textwidth, bb=0 0 900 300]{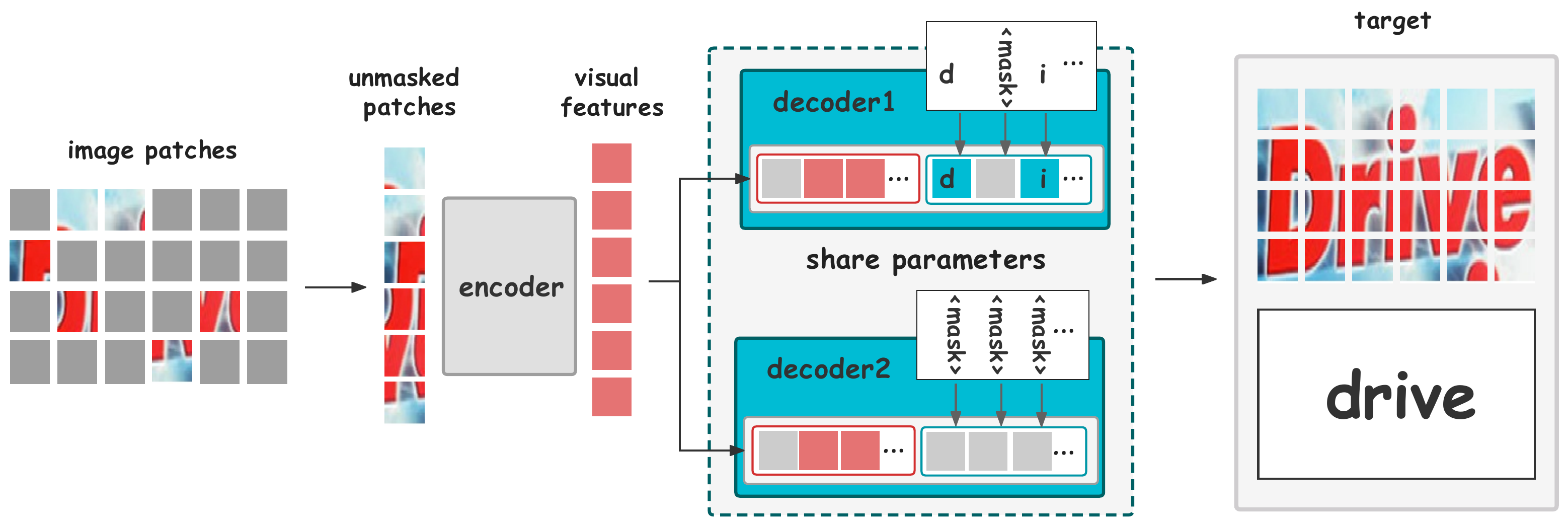}
	\caption{An illustration of MVLT in the pretraining stage.}
	\label{framework_1}
\end{figure}

\label{sec:method}
Our proposed MVLT is trained with two stages, including a pretraining stage, as shown in Figure~\ref{framework_1}, and a fine-tuning stage, as shown in Figure~\ref{framework_2} (a). In addition, we harness the unlabeled data to enhance the training of the model, as shown in Figure~\ref{framework_2} (b) and (c). 

\subsection{Preliminary}

{\bf Vision Transformer (ViT).} To model 2D image into Transformer, which is designed to process 1D sequence, ViT \cite{DBLP:conf/iclr/DosovitskiyB0WZ21}  flattens an image by reshaping the image $x \in \mathbb{R}^{H \times W \times C}$ into a sequence of image patches $x \in \mathbb{R}^{N \times (P^{2} \cdot C)}$, where $(H,W)$ denotes the original scale of the image, $C$ is the number of channels, and  $(P,P)$ is the resolution of each image patch. 


{\bf Masked autoencoders (MAE).} MAE \cite{he2022masked}
randomly divides $N$ image patches into $N_u$ unmasked ones and $N_m$ masked ones. Correspondingly, $x=[x_u;x_m] \in \mathbb{R}^{N \times (P^{2} \cdot C)}$, where $x$ is the full set of image patches, $x_u$ is the set of unmasked patches, and $x_m$ is the set of masked patches. The encoder of MAE is a ViT, which only operates on $x_u$  to learn the visual feature embeddings:
 \begin{align}
 \label{mae_encoder}
      v_u=\text{encoder}(x_u),
 \end{align}
 where $v_u \in \mathbb{R}^{N_u \times D_1}$ and $D_1$ is the feature dimension in the encoder. 
 The mask token $v_m$ is introduced after the encoder, constituting the input of the decoder, together with $v_u$. The decoder reconstructs the image as follows:
 
 \begin{align}
 \label{mae_decoder}
      \hat{v}_m,\hat{v}_u=\text{decoder}(v_m,v_u),
 \end{align}
where $v_m \in \mathbb{R}^{N_u \times D_2}$ and $D_2$ is the feature dimension in the decoder.
$\hat{v}_m \in \mathbb{R}^{N_m \times (P^{2} \cdot C)}$ and $\hat{v}_u \in \mathbb{R}^{N_u \times (P^{2} \cdot C) }$. The Mean Squared Error (MSE) loss is used to optimize MAE model.

\subsection{Pretraining Stage}
In pretraining, MVLT intends to reconstruct the masked image patches and recognize the scene text from the masked image. Like MAE, the reconstruction of image patches helps our model to learn an effective visual representation. However, recognizing texts is beyond the scope of MAE, and we propose a novel language-aware model to deal with it, as shown in Figure~\ref{framework_1}.

{\bf Masked Encoder.}  We use ViT as the encoder of MVLT, which is same as Eq.~(\ref{mae_encoder}). Each image is split into $N$ patches. With a mask ratio of 0.75, we divide the patches into a set of masked ones $x_m \in \mathbb{R}^{N_m \times (P^{2} \cdot C)}$, and a set of the unmasked ones $x_u \in \mathbb{R}^{N_u \times (P^{2} \cdot C)}$. The encoder embeds $x_{u}$ by using it as the input of a linear layer and then adding the positional embeddings with the output of the linear layer. Next, the encoder uses a series of Transformer blocks to learn the visual embedding: $ v_u=\text{encoder}(x_u)$.

{\bf Masked Decoder.}
We build a multi-modal Transformer as the decoder to exploit the visual cues and linguistic information in each image. The visual cues come from $v_{u}$.
The linguistic information comes from the word-level text label that is character-wise mapped into a sequence of learnable character embeddings $t \in \mathbb{R}^{L \times D_2}$, where $L$ is the length of the character embedding sequence.
Similar to the image patches, we denote $t=[t_u, t_m]$, where $t_u \in \mathbb{R}^{L_u \times D_2}$ is the unmasked character embeddings, and $t_m \in \mathbb{R}^{L_m \times D_2}$ is a sequence of the special "<mask>" token embeddings, $L_u$ and $L_m$ are the corresponding length. We add positional embeddings to ($v_{m}$, $v_{u}$, $t_m$, $t_u$) to build the decoder input. For symbolic simplification, we keep the symbols before and after adding the positional embeddings the same. We design two sub-decoders ($decoder_1$ and $decoder_2$), applying different masking strategies on $t$ to learn textual semantics. Note that the two sub-decoders share parameters during pretraining. Our decoder is denoted as:
\begin{align}
    \hat{v}_m,\hat{v}_u,\hat{t}_m,\hat{t}_u=\text{decoder}(v_m,v_u,t_m,t_u),
\end{align}
where $\hat{v}_m \in \mathbb{R}^{N_m \times (P^{2} \cdot C)}$, $\hat{v}_u \in \mathbb{R}^{N_u \times (P^{2} \cdot C)}$, $\hat{t}_m \in \mathbb{R}^{L_m \times M}$, $\hat{t}_u \in \mathbb{R}^{L_u \times M}$, and $M$ is the number of character's category.

 \textit{ Modeling explicit  language semantics.}  We set the text mask ratio of $decoder_1$ to 0.2. For instance, if the text label consists of 10 characters, we randomly mask 2 of them.  The length of unmasked character embeddings $L_u=8$, and the length of masked ones $L_m=L-L_u$. With the unmasked character embeddings $t_u$ serving as linguistic context, $decoder_1$ explicitly learns textual semantics.

 \textit{ Modeling implicit language semantics.}    For $decoder_2$, we set the mask ratio of the character embeddings as 1.0. With this setting, the length of unmasked character embeddings $L_u=0$, and the masked ones $L_m=L$.  As the characters are totally masked, $decoder_2$ only uses the visual information to reconstruct $\hat{v}_{m}$ and predicts the word-level text label $\hat{t}_m$, which pushes it to learn implicit textual semantics from visual cues.

{\bf Pretraining objective.}
We use the MSE loss to optimize the reconstructed image patches and use the Cross Entropy Loss to optimize textual prediction:
\begin{align}
\label{pre}
    \mathcal{L}_{\text{pretraining}}=
    \alpha \cdot \mathcal{L}_{v_1} +
    \beta \cdot \mathcal{L}_{v_2} + 
    \gamma \cdot \mathcal{L}_{t_1} + 
    \epsilon \cdot \mathcal{L}_{t_2},
\end{align}
 where \{$\alpha$,$\beta$,$\gamma$, $\epsilon$\} are trade-off parameters. For $decoder_1$, $\mathcal{L}_{v_1}=\text{MSE}(\hat{v}_{m},y_m)$, where $y_m \in \mathbb{R}^{N_m \times (P^{2} \cdot C)}$ denotes the pixel values of the masked image patches. $\mathcal{L}_{t_1}=\text{Cross-Entropy}(\hat{t}_{m},y_t)$, where $y_t \in \mathbb{R}^{L \times 1}$ is the ground-truth character index, and each element in $y_t \in [1,M]$. For $decoder_2$, $\mathcal{L}_{v_2}=\text{MSE}(\hat{v}_{m},y_m)$, and $\mathcal{L}_{t_2}=\text{Cross-Entropy}(\hat{t}_{m},y_t)$.
 
 \begin{figure}[t]
    \centering
	\includegraphics[width=0.97\textwidth, bb=0 0 1100 500]{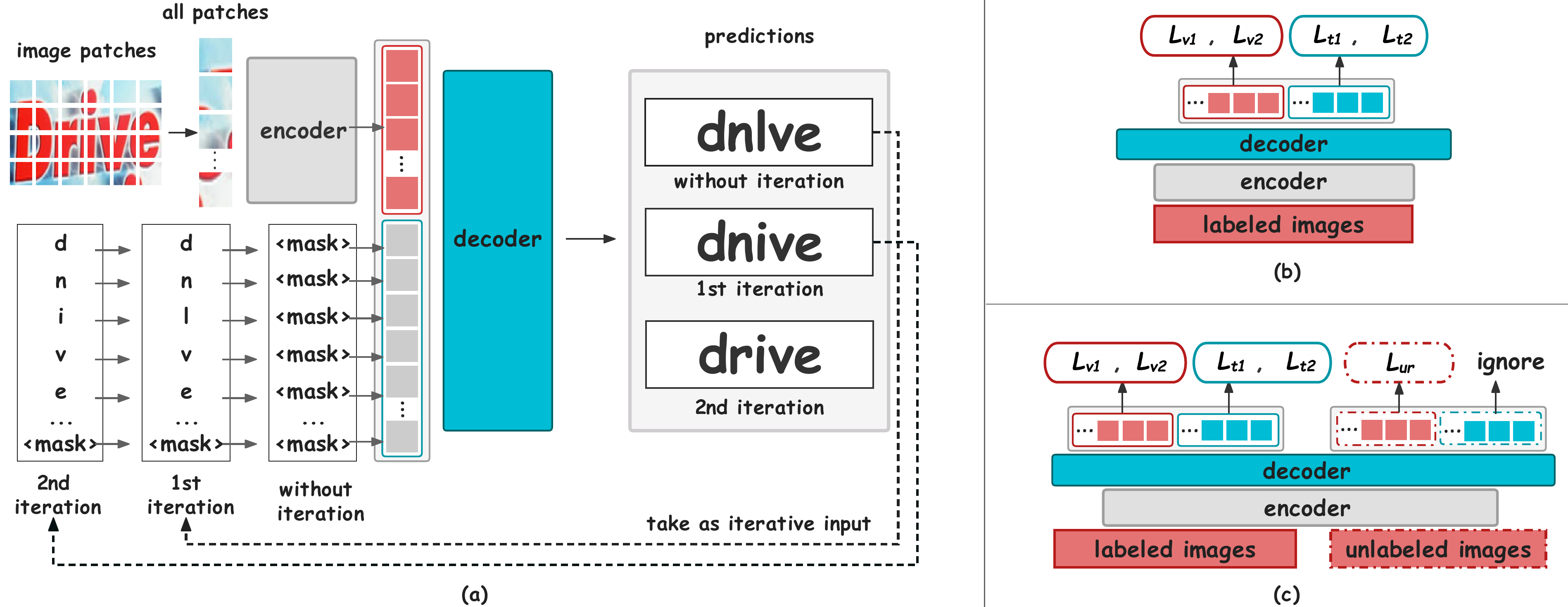}
	\caption{Sub-figure (a) shows our model in the fine-tuning stage. Sub-figure (b) illustrates our model using labeled data in the pretraining stage. Sub-figure (c) illustrates our model  using both labeled and unlabeled data in the pretraining stage.}
	\label{framework_2}
\end{figure}

 \subsection{ Fine-tuning Stage}
 We fine-tune the above pre-trained encoder and decoder to further promote the performance of our model on the STR task. Motivated by the iterative strategy of ABINet~\cite{DBLP:conf/cvpr/FangXWM021}, we design an iterative correction method that fits the architecture of our model, as shown in Figure \ref{framework_2} (a). This iterative correction method is an alternative during fine-tuning.

 {\bf Encoder.} The full set of image patches $x \in \mathbb{R}^{N \times (P^{2} \cdot C)}$ is taken as the input of the encoder without utilizing the masking operation:
\begin{align}
    \label{q5}
   v&=\text{encoder}(x),
   \end{align}
where $v \in \mathbb{R}^{N \times D_1}$ and $N$ is the number of image patches.

{\bf Decoder.} In the pretraining stage, our encoder takes the output of the encoder (visual-related feature) and a sequence of character embeddings (linguistic-related feature) as input. However, in the fine-tuning stage, when the iterative correction method is not used or before the first iteration, only the visual-related feature is visible by the decoder, without the linguistic-related feature. Thus, to be consistent with the input of the pretrained decoder, we use a sequence of "<mask>" token embeddings as the character embedding, $t_m \in \mathbb{R}^{L \times D_2}$. Different from the pretraining stage, where the output of the decoder consists of reconstructed image patches and the predicted text, during the fine-tuning stage, the predicted image patches are ignored, and only the predicted text is kept because the STR task only focuses on the text prediction. The input and output of the decoder are formalized as follows:
  \begin{align}
      \label{fine-decoder}
      \hat{t}=\text{decoder}(v,t_m),
\end{align}
where $\hat{t} \in \mathbb{R}^{L \times M}$ is the logits of predicted text. 

\textbf{\textit{Iterative correction.}}
We regard the output $\hat{t}$ as a raw text prediction that will be corrected during iterations. The prediction probability of each character is computed based on $\hat{t}$. 
A linear projection layer takes the prediction probabilities as the input. It outputs a sequence of character-related feature representations, which is set as the new character embedding input at the current iteration. For example,
we perform $K$ times of iterative corrections by the follows:
for the 1st iteration, we pass $\hat{t}_\text{itr=0}=\hat{t}$ into a softmax layer and then a linear layer to get $t_{\text{itr=1}}$, which is taken as the new character embedding to replace the totally masked character embeddings $t_m$ in Eq.~(\ref{fine-decoder}). 
 Then, the k-th ($k \in [1, K]$) iteration process is formalized as:
\begin{align}
    \text{prob}_\text{itr=k}&=\text{softmax}(\hat{t}_\text{itr=k-1}), \\
    t_\text{itr=k}&=\text{linear}( \text{prob}_\text{itr=k}),\\
    \hat{t}_\text{itr=k}&=\text{decoder}(v,t_\text{itr=k}),
    \label{iterative-decoder}
\end{align}
The output of the K-th iteration is regarded as the final corrected text prediction.

{\bf Fine-tuning objective.} Different from the pretraining objective, we only focus on optimizing the textual prediction by leveraging the Cross Entropy Loss:
\begin{align}
    \mathcal{L}_{\text{fine-tuning}}= \frac{1}{2}\text{Cross-Entropy}(\hat{t}_{\text{itr=0}},y_t) +         \frac{1}{2(K-1)}\sum_{j=1}^{K}\text{Cross-Entropy}(\hat{t}_{\text{itr=j}},y_t).
\end{align}

\subsection{Using Unlabeled Real Dataset}

Our pretraining objective includes: (a) reconstructing the masked image patches; and (b) recognizing the scene texts. 
Since (b) requires leveraging labeled scene text to attain supervised training, 
we use labeled synthetic datasets for training.
However, there is a mass of real-world data, which is unlabeled and costs much to label. Fortunately, we find it easy to  use unlabeled data to enhance learning features during pretraining. 
Specifically, as shown in Figure \ref{framework_2} (c), we concatenate the unlabeled image data and the labeled image data along the batch dimension as a new input batch. For a batch of output $O= \{O_{syn},O_{ur}\}$,  ${O}_{syn} = \{ (\hat{v}_m,\hat{v}_u,\hat{t}_m,\hat{t}_u)_1, ...,  (\hat{v}_m,\hat{v}_u,\hat{t}_m,\hat{t}_u)_{N_{1}}\}$ is the output corresponding to the labeled synthetic datasets, and $O_{ur}= \{(\hat{v'}_m,\hat{v'}_u,\hat{t'}_{m},\hat{t'}_{u})_1, ..., (\hat{v'}_m,\hat{v'}_u, \hat{t'}_m, \hat{t'}_u)_{N_{2}}\}$ is the output corresponding to the unlabeled real datasets, where $N_{1}$ is the number of labeled synthetic data in the batch,  and $N_{2}$ is the number of unlabeled real data in the batch. $O_{syn}$ is optimized on the loss function of Eq.~(\ref{pre}), while $O_{ur}$ is optimized only on a MSE loss $\mathcal{L}_{ur}=\text{MSE}(\hat{v'}_m,y'_{m})$, $\hat{v'}_m$ is the output corresponding to the masked patches of the real data, and $y'_{m}$ is the target pixels of the real data. Through this semi-supervised pretraining process, we can learn real-world domain knowledge, thus reducing the gap between the pretraining data and the real-world scenario.  

\section{Experiment}
\label{Experiments}

{\bf Datasets.} To conduct supervised training, we use two synthetic datasets,  MJSynth (MJ)~\cite{jaderberg2016reading,Jaderberg14c} and SynthText (ST)~\cite{DBLP:conf/cvpr/GuptaVZ16}, as the labeled training dataset of MVLT. We use all of the 14 real datasets that collected in~\cite{DBLP:conf/cvpr/BaekMA21}, and remove all the image labels to build our unlabeled real dataset. We denote the unlabeled real dataset as UR for a simplified description. 
We use the same test dataset as ABINet \cite{DBLP:conf/cvpr/FangXWM021}, including six standard benchmarks, which consists of 857 images from ICDAR2013 (IC13) \cite{DBLP:conf/icdar/KaratzasSUIBMMMAH13}, 1,811 images from  ICDAR2015 (IC15) \cite{DBLP:conf/icdar/KaratzasGNGBIMN15}, 647 images from  Street View Text (SVT) \cite{DBLP:conf/iccv/WangBB11}, 645 images from SVT-Perspective (SVTP) \cite{DBLP:conf/iccv/PhanSTT13}, 3,000 images from IIIT 5-K Words (IIIT) \cite{DBLP:conf/bmvc/MishraAJ12}, and 288 images from CUTE80 (CUTE) \cite{DBLP:journals/eswa/RisnumawanSCT14}.

\textbf{Model detail.} Images are scaled to 112$\times$448, with a resolution of 14$\times$14 for each patch. Although the image size is different from ViT \cite{DBLP:conf/iclr/DosovitskiyB0WZ21}, we keep the number of patches the same. Our encoder uses the same settings as ViT-B in MAE \cite{he2022masked}. We use a lightweight decoder, which has depth 4, width 512, and 8 attention heads. Thus the dimension $D_1$ and $D_2$ are set to 768 and 512, respectively. The length of the character embedding sequence, $L$, is set to 27, because, according to our observation, the vast majority of the word is shorter than 27 characters. In the pretraining stage, to build the input of $decoder_1$, we pad the text labels that are shorter than $L$ by a special token "<mask>". The input of $decoder_2$ is built from $L$ "<mask>" tokens.

\textbf{Training detail.} The trade-off parameters $\alpha$ and $\beta$ are set to 0.5, $\gamma$ and $\epsilon$ are set to 0.01.In both the pretraining and the fine-tuning stage, an AdamW \cite{loshchilov2018decoupled} optimizer and a cosine learning rate decay scheduler are applied. In the pretraining stage, we set the initial learning rate to 1.5e-4 and weight decay to 0.05. When using only the labeled data, the batch size is set to 4,096.  When using both the labeled and the unlabeled data, the batch size is 6,144, with 4,096 labeled images and 2,048 unlabeled ones. We conduct a total of 120,000 iterations, with 8,000 warm-up iterations. The optimizer momentum is set to $\beta_1$=0.9 and $\beta_2$=0.95. We do not use grad clip during pretraining. In the fine-tuning stage, the batch size is  1,024, the initial learning rate is 1e-5, and the weight decay is 0.05. We conduct a total of 20,000 iterations, with 8,000 warm-up iterations.The optimizer momentum is set to $\beta_1$=0.9 and $\beta_2$=0.999. The grad clip is set to 2.0. The layer-wise learning rate decay is set to  0.75. We perform 3 times of iterative corrections in fine-tuning the model and 3 times of iterative corrections in testing the model. We use 8 NVIDIA RTX A6000 GPUs, with 48GB memory, to conduct the experiments, and use gradient accumulation to maintain a large effective batch size. The pretraining stage takes around 3.5 days, and the fine-tuning stage takes around 5 hours. 

\textbf{Data augmentation.} In the pretraining stage, we apply RandomResizedCrop to augment data, which is similar to MAE. Specifically, we set the scale to (0.85, 1.0) and the ratio to (3.5, 5.0). In the fine-tuning stage, we use the same data augmentation method as ABINet, including  rotation, affine, and perspective.

\begin{table*}
	\centering
	\footnotesize
	\caption{Accuracy results of our MVLT and SOTA methods on six regular and irregular STR datasets. "UR" denotes the unlabeled real dataset.}
	\label{dataset_accuracy}
    \begin{tabular}{c|c|ccc|ccc}
		\toprule
		\multirow{2}{*}{Method}&	\multirow{2}{*}{Datasets} & 	
		\multicolumn{3}{c|}{Regular Text} 	& \multicolumn{3}{c}{Irregular Text} 	
    
		\cr
	&&IC13&SVT&IIIT&IC15&SVTP&CUTE\cr
		\midrule
	
		ASTER~\cite{DBLP:journals/pami/ShiYWLYB19} &MJ+ST&91.8&89.5&93.4&76.1&78.5&79.5 \cr
			ESIR~\cite{DBLP:conf/cvpr/ZhanL19}&MJ+ST&91.3&90.2&93.3&76.9&79.6& 83.3\cr
		ScRN~\cite{DBLP:conf/iccv/YangGLHBBYB19} &MJ+ST&93.9&88.9&94.4& 78.7&80.8&87.5\cr
			PIMNet~\cite{DBLP:conf/mm/QiaoZWWZJWW21} &MJ+ST&93.4&91.2&95.2&81.0&84.3&84.4   \cr
	SAR~\cite{DBLP:conf/aaai/00130SZ19}&MJ+ST&94.0&91.2&95.0&78.8&86.4&89.6 \cr
	SRN~\cite{DBLP:conf/cvpr/YuLZLHLD20} &MJ+ST&95.5&91.5&94.8& 82.7&85.1&87.8\cr
	GTC~\cite{DBLP:conf/aaai/HuCHYL20}&MJ+ST&94.3&92.9&95.5&82.5&86.2&92.3\cr
		
	VisionLAN~\cite{DBLP:conf/iccv/WangXFWZZ21} &MJ+ST&95.7&91.7&95.8&83.7&90.7&88.5 \cr
	PREN2D~\cite{DBLP:conf/cvpr/YanPXY21} &MJ+ST&96.4&94.0&95.6& 83.0&87.6&91.7\cr
	S-GTR~\cite{DBLP:conf/aaai/HeC0LHWD22} &MJ+ST&96.8&94.1&95.8& 84.6&87.9&92.3\cr
	ABINet~\cite{DBLP:conf/cvpr/FangXWM021} &MJ+ST&97.4&93.5&96.2& 86.0&89.3&89.2\cr

   	\midrule
 \textbf{MVLT}  &MJ+ST& 97.3 &94.7 &96.8 &87.2 &90.9 &91.3 \cr
 \textbf{MVLT}$^{*}$  &MJ+ST+UR&  \textbf{98.0}& \textbf{96.3}& \textbf{97.4}& \textbf{89.0} & \textbf{92.7}& \textbf{95.8}\cr
		\bottomrule
	\end{tabular}

\end{table*}

\subsection{Performance Analysis}
{\bf Comparisons with state-of-the-arts.} Table~\ref{dataset_accuracy} summarizes the comparison result of our proposed method with eleven SOTA. MVLT is trained on ST and MJ, and MVLT* uses extra unlabeled data. MVLT achieves the highest accuracy in most of the datasets. The result proves the effectiveness of our design of model architecture and our two-stage training procedure. Benefiting from learning linguistic knowledge, MVLT is more tolerant to irregular texts, leading to a higher performance even compared with the methods that use rectification modules, such as ASTER, ESIR, and GTC. Compared with other language-aware models, including VisionLAN, ABINet, PREN2D, etc., our model performs better on the vast majority of the test datasets, further justifying the capture of textual semantics. To perform a fair comparison, MVLT uses the same data and the data augmentation method as ABINet in the fine-tuning stage. MVLT outperforms ABINet with 1.2\%, 0.6\%, 1.2\%, 1.6\%, and 2.1\% on SVT, IIIT, IC15, SVTP, and CUTE datasets, respectively. 

 {\bf Using unlabeled real dataset.} 
As shown in Table \ref{dataset_accuracy}, MVLT* outperforms MVLT on most  test datasets. Specifically, MVLT* largely promotes the result of recognizing irregular texts, showing that using unlabeled real data is able to enhance the ability of STR,  leading to a more practicable model facing  the real-world scenario.

\begin{figure}
	\centering
	\subfigure[Without unlabeled datasets.]{
		\centering
		\includegraphics[width=0.28\textwidth, bb = 0 0 300 400]{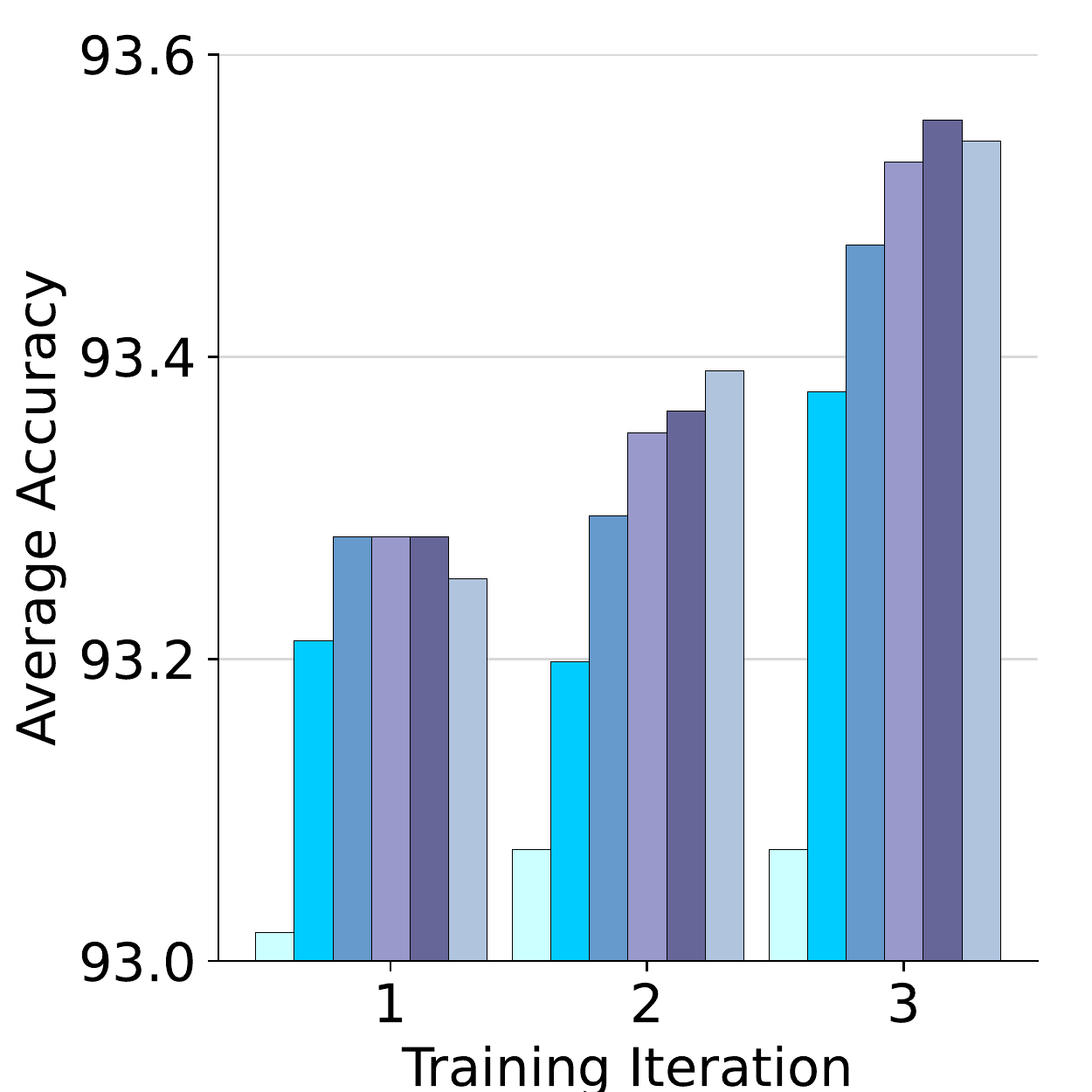}
	}\hspace{10mm}
	\subfigure[With unlabeled datasets.]{
		\centering
		\includegraphics[width=0.4\textwidth, bb = 0 0 430 400]{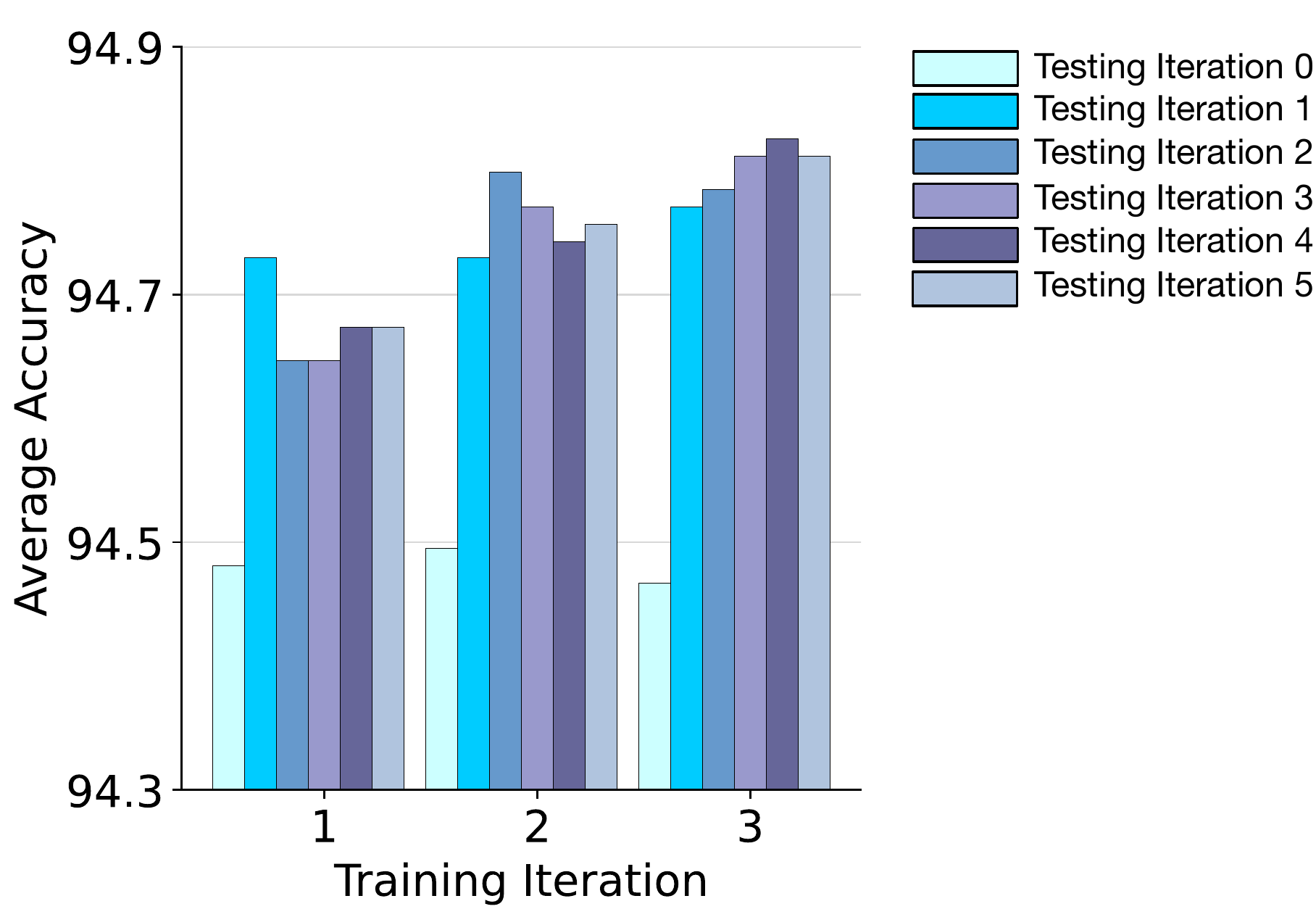}
	}
	\caption{Accuracy of iterative correction in the training and testing process.}
	\label{exp-efficiency}
\end{figure}
{\bf Iterative correction.}
Iterative correction is applied during training the model in the fine-tuning stage, as well as testing. Figure~\ref{exp-efficiency} shows the results with different settings of the iterate times, which suggests that the iterative correction method can help obtain higher accuracy.

\subsection{Ablation Study}
\begin{table} [htbp]
	\centering
	\footnotesize
	\caption{The results of ablation study. "Iter" represents using iterative correction method in training the model in the fine-tuning stage, and in testing on the test datasets.}
	\label{ablation}
    \begin{tabular}{|ccccc|ccc|ccc|c|}
		\toprule
	        	\multirow{2}{*}{$\mathcal{L}_{v_1}$} &
 	             \multirow{2}{*}{$\mathcal{L}_{t_1}$}&
 	             \multirow{2}{*}{$\mathcal{L}_{v_2}$} & \multirow{2}{*}{$\mathcal{L}_{t_2}$}& 
 	              \multirow{2}{*}{Iter}&
 	             \multicolumn{3}{c|}{Regular Text} &    \multicolumn{3}{c|}{Irregular Text} &    \multirow{2}{*}{Total} \cr &&&&&IC13&SVT&IIIT&IC15&SVTP&CUTE& \cr
		\midrule
$\surd$& $\surd$&$\surd$&$\surd$&$\surd$& 97.3 &94.7 &96.8 &87.2 &90.9 &91.3 &93.5 \cr
$\surd$& $\surd$&$\surd$&$\surd$&&97.0 &94.1 &96.8 	&86.6 &89.6 &90.6 &93.2 \cr
&&$\surd$&$\surd$&$\surd$&96.8 &94.4 &96.3 &87.0 &	89.9 &90.6 &93.1 \cr
&&$\surd$& $\surd$&& 96.7 &94.6 &96.3 &86.7 &90.2 &91.0 &93.1 \cr
$\surd$& $\surd$&&&$\surd$& 96.4 &92.4 &96.4 &86.7 &89.3 &91.0 &92.8 \cr
$\surd$& $\surd$&&&& 95.9 &92.7 &96.0 &86.3 &88.7 &	89.6 &92.4 \cr
$\surd$& & & & & 96.4 & 91.3& 96.5& 85.5& 88.8& 89.9& 92.3  \cr
		\bottomrule
	\end{tabular}
\end{table}

The statistics in Table \ref{ablation} show that: 1) Using only the visual-related loss $\mathcal{L}_{v_1}$ is not as effective as using both the visual-related and the textual-related losses. 2) Losing either the loss related to learning explicit semantics or the loss related to learning implicit semantics will result in a decrease in accuracy. 3) Learning the implicit textual semantics leads to a higher leap in accuracy than learning the explicit semantics. 4) The iterative correction is especially useful when the model is trained with explicit textual semantics.

\subsection{Visualization and Analysis}
\label{visualization}

\begin{figure}[h]
    \centering
	\includegraphics[width=0.98\textwidth, bb=0 0 830 200]{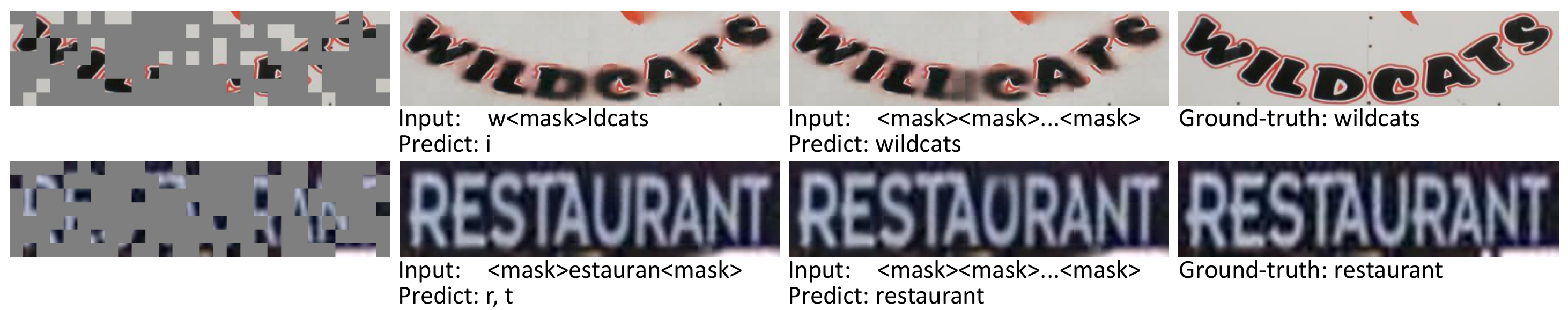}
	\caption{Visualization of pretrained MVLT. For each example, we show the masked image (left), our image reconstruction and text prediction from $decoder_1$ (mid-left) and $decoder_2$ (mid-right), and the ground-truth (right).}
	\label{visualize1}
\end{figure}

\begin{figure}[h]
    \centering
	\includegraphics[width=0.98\textwidth, bb = 0 0 830 300]{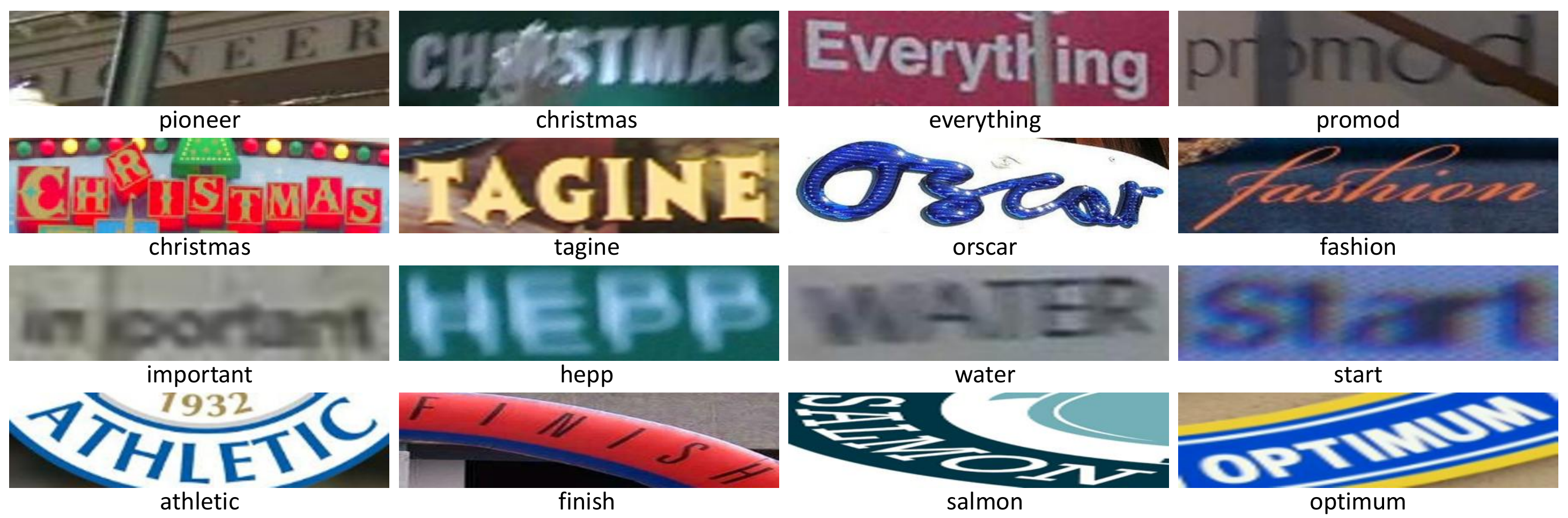}
	\caption{Visualization of fine-tuned MVLT. We show successfully recognized difficult examples with occlusion (1st row), complex font styles (2nd row), blur (3rd row), and deformation (4th row).}
	\label{visualize2}
\end{figure}

Figure \ref{visualize1} and Figure \ref{visualize2} display several visualization results after the pretraining and fine-tuning stage. The images are from test datasets. Figure \ref{visualize1} suggests that the model has already acquired visual and linguistic semantic knowledge through the pretraining stage. Figure~\ref{visualize2} shows a stronger model after fine-tuning, which is capable of dealing with more challenging real-world images. More examples are shown in the supplementary material.

\section{Conclusion}
\label{Conclusion}
We propose a Masked Vision-Language Transformers (MVLT) for STR, getting superior results compared to the state-of-the-art models. For pretraining the model, we design a masking strategy to lead the model in learning both the explicit and implicit textual semantics. Experiment results have proved the effectiveness of our model in capturing semantics and the usefulness of both kinds of semantics. During training and testing our model in the fine-tuning stage, we apply an iterative correction method, which boosts the performance of STR. Furthermore, the use of unlabeled real data improves the applicability of our model in real-world scenarios.

\bibliography{bmvc_final}

\clearpage
\center\textbf{Supplementary Material}
\begin{figure}[!h]
    \centering
	\includegraphics[width=1
	\textwidth, bb=0 0 1700 2200]{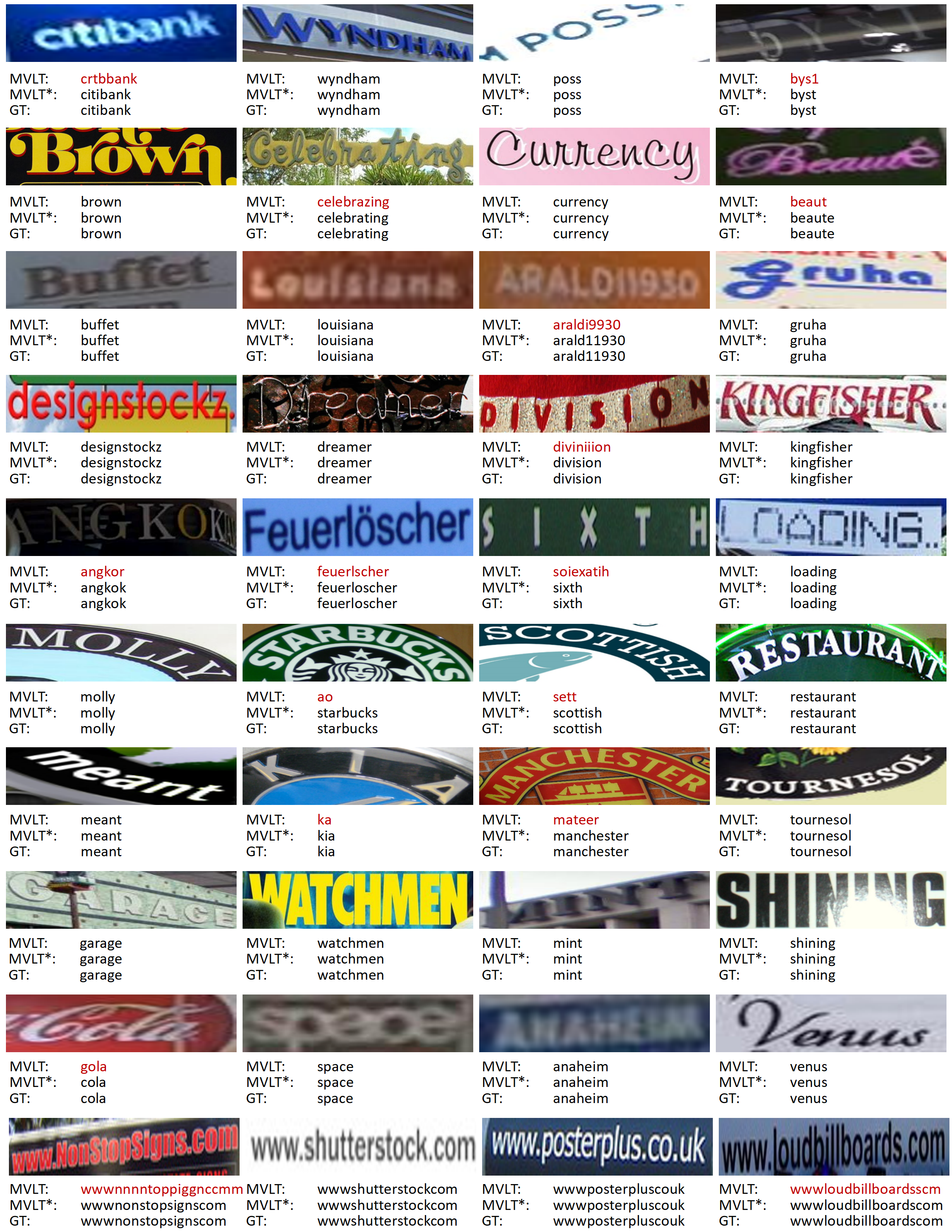}
	\caption{Visualization of fine-tuned MVLT and MVLT*. For each example, we show the STR results of MVLT, MVLT*, and the ground-truth. The wrong predictions are shown in red.  The images are from test datasets.}
\end{figure}

\begin{figure}[ht]
    \centering
	\includegraphics[width=1
	\textwidth, bb = 0 0 1700 2200]{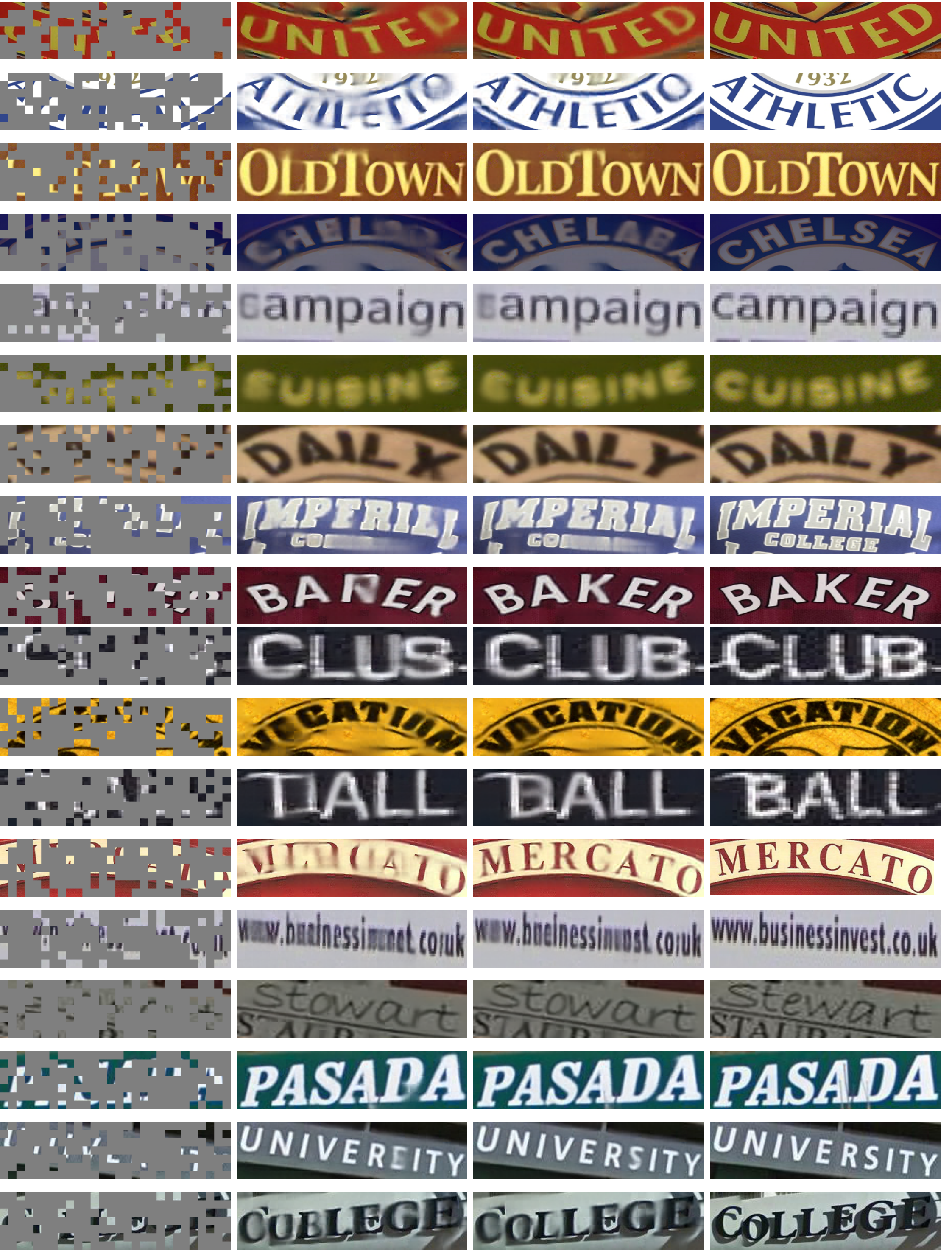}	\caption{Visualization of pretrained MVLT and MVLT*. For each example, we show the masked image (left), the image reconstruction result of $decoder_2$ of the pretrained MVLT (mid-left), the image reconstruction result of $decoder_2$ of the pretrained MVLT* (mid-right), and the ground truth (right). The images are from test datasets.}
\end{figure}

\end{document}